\documentclass{article}
\usepackage{spconf,amsmath,amssymb,graphicx,booktabs,cite,xcolor,url}
\usepackage{pgfplots}
\pgfplotsset{compat=1.18}
\usepgfplotslibrary{groupplots}
\usepackage[table]{xcolor}
\usepackage{multirow}
\definecolor{bestcyan}{RGB}{225,245,250}

\title{FD-VAD: SEMANTIC ENDPOINT DETECTION FOR STREAMING FULL-DUPLEX SPEECH}

\name{Puneet Mathur, Dinesh Manocha}
\address{University of Maryland, College Park}

\begin{document}
\ninept
\maketitle

\begin{abstract}
Natural turn-taking in full-duplex voice interaction requires determining from partial speech whether a pause reflects hesitation or a completed conversational intent. Acoustic voice activity detection lacks this semantic information, while cascaded ASR-based endpointing introduces transcription dependence and additional processing stages. We formulate semantic endpoint detection as a \emph{causal audio-language reasoning task} and introduce \textbf{FD-VAD}, an ASR-free streaming endpointer that maps bounded causal audio windows directly to \textsc{Continue}/\textsc{Stop} decisions. FD-VAD combines a frozen speech encoder with a lightweight modality adapter and a parameter-efficiently adapted language model, using a last-chunk training objective for streaming inference. We further introduce confidence-gated endpoint commitment to control interruption versus delay and boundary-focused hard-negative sampling to improve decisions around ambiguous turn boundaries. Across in-domain and conversational evaluations, FD-VAD outperforms strong streaming and non-streaming semantic turn classifiers, and achieves the highest EOT recall among qualifying systems on TurnBench dev set $0.853$ (at $\mathrm{FP}\leq0.10$) in a zero-shot setting. These results show that semantic endpointing can be performed directly from streaming audio without intermediate ASR or dialogue state tracking.
\end{abstract}


\section{Introduction}

Natural spoken interaction depends not only on \emph{what} is said, but also on \emph{when} each participant chooses to speak. Humans continuously infer whether another speaker has completed a thought, is pausing to plan what to say next, or intends to continue after a hesitation. This ability is fundamental to fluid turn-taking: responding before a speaker has finished is disruptive, while waiting too long after a completed request makes an interaction feel unnatural. The problem becomes particularly important for full-duplex voice agents, which continuously listen and speak, and therefore cannot rely on the rigid listen--respond alternation used by conventional voice assistants~\cite{moshi,freezeomni}. A full-duplex agent must instead maintain an evolving estimate of whether the user still holds the conversational floor.

Voice activity detection (VAD) is commonly used in spoken systems to locate acoustic speech boundaries. However, acoustic VAD is often insufficient because a period of silence does not necessarily indicate that a conversational turn has ended. A speaker may pause while searching for a word, producing a disfluency, or planning the remainder of a request; conversely, a short utterance may already express a complete intent. The decision is therefore inherently \emph{semantic}: the system must determine whether the observed speech constitutes a complete conversational act, rather than merely detect whether acoustic energy has disappeared. We refer to this problem as \emph{semantic endpoint detection}. In a streaming setting, it is especially challenging because the decision must be updated from partial audio without access to future speech.

\begin{figure}[t]
    \centering
    \includegraphics[width=\columnwidth]{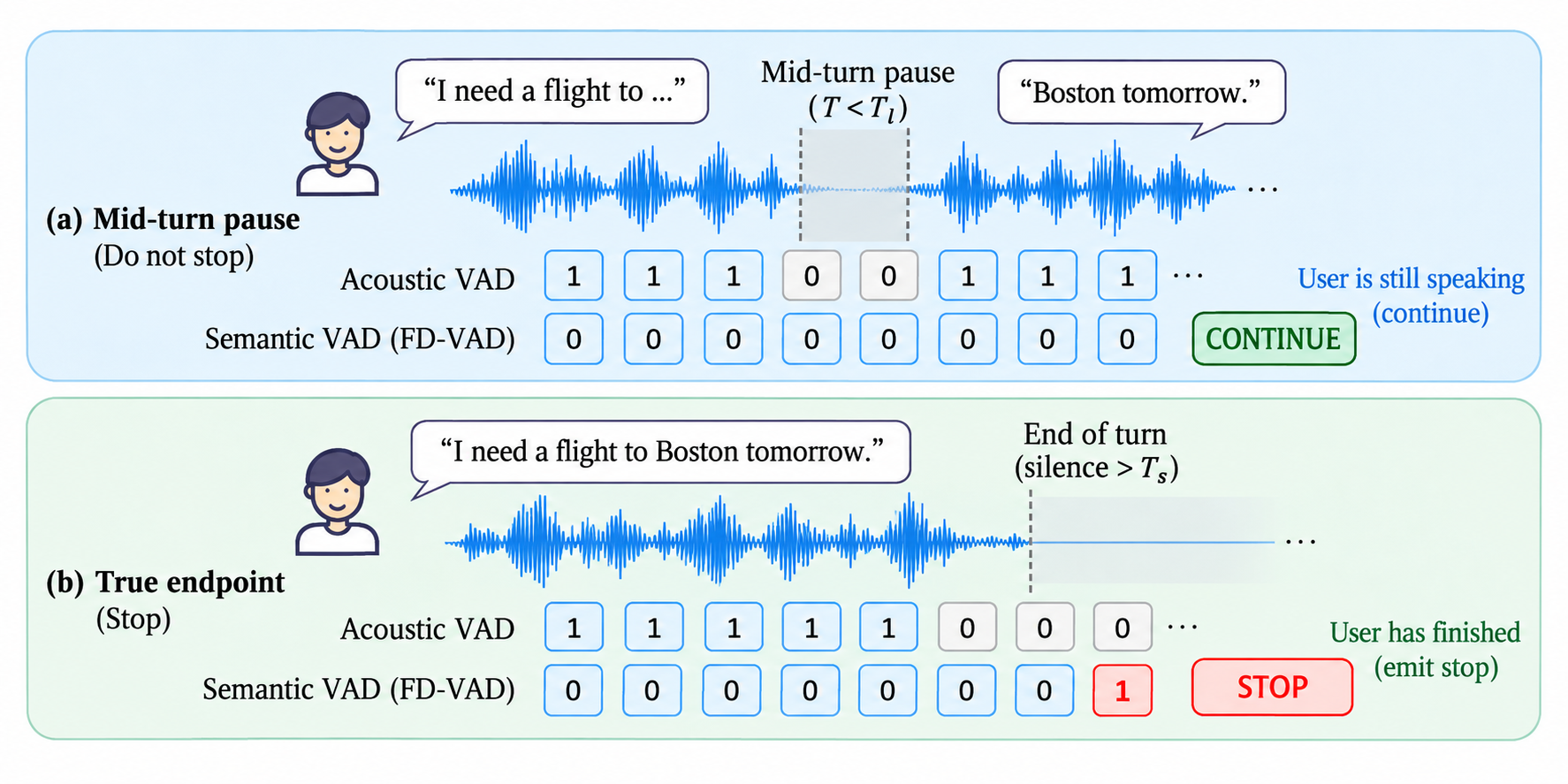}
    \caption{\small{Streaming semantic endpoint detection task, where 0 and 1 denote \textsc{Continue} and \textsc{Stop}, respectively. \textbf{FD-VAD} uses linguistic as well as acoustic cues for end-of-turn detection.}}
    \label{fig:semantic_vad_intro}
\end{figure}

Existing endpointing approaches expose a trade-off between semantic reasoning, streaming operation, and architectural simplicity. Acoustic methods such as Silero VAD~\cite{silero} are lightweight and naturally streaming but cannot determine whether the preceding speech is semantically complete. Cascaded approaches first transcribe speech and then use a language model or text classifier to determine whether the utterance is complete~\cite{ten,paraformer}. These methods provide access to lexical semantics, but introduce an explicit ASR dependency into the turn-taking loop and inherit transcription errors and additional processing stages. Semantic approaches such as Smart Turn~\cite{smartturn}, Easy Turn~\cite{easyturn}, FastTurn~\cite{fastturn}, Next-Turn~\cite{nextturn}, and SpeculativeETD~\cite{speculativeetd} incorporate semantics through utterance-level classification, transcript/CTC pathways, timing supervision, or silence-triggered semantic inference. At the other end, full-duplex systems such as Moshi and Freeze-Omni~\cite{moshi,freezeomni} learn turn-taking jointly with response generation. These methods improve semantic awareness, but typically introduce intermediate dependencies, triggering mechanisms or couple to particular dialogue architectures.

We argue that semantic endpointing need not require these intermediate mechanisms and instead formulate it as a \emph{streaming audio-language reasoning task}: given only the speech observed so far, determine whether the speaker has expressed a complete conversational intent. We introduce \textbf{FD-VAD}, an ASR-free semantic endpoint detector for streaming full-duplex speech. FD-VAD maps short causal waveform windows directly to \textsc{Continue}/\textsc{Stop} decisions. We utilize a frozen self-supervised speech encoder extracts acoustic and linguistic representations from the incoming waveform, while a lightweight modality adapter maps them into the embedding space of a small language model that performs the semantic endpoint decision. Rather than processing the complete utterance, FD-VAD repeatedly evaluates a bounded sliding window and supervises only the state at its most recent chunk. The resulting model reasons about semantic completeness while preserving causal, fixed-cost streaming inference. Importantly, FD-VAD requires neither transcript generation nor a separate acoustic-to-semantic cascade, and its endpointing policy remains independent of the downstream voice agents.

Additionally, we address two properties particularly important for real-time deployment. First, endpoint errors are asymmetric: an early \textsc{Stop} can interrupt a user, whereas a conservative decision primarily adds delay. We therefore introduce a \textbf{confidence-gated decision mechanism} that explicitly controls this trade-off through the predicted stop probability and the number of consecutive decisions required before committing an endpoint. Second, ambiguous examples are concentrated near the true turn boundary rather than uniformly throughout an utterance. To better model these cases, we introduce \textbf{boundary-focused hard-negative sampling}, which emphasizes difficult boundary regions during training while leaving the inference architecture unchanged. \textbf{Our main contributions are:}

\begin{itemize}
    \item We reformulate semantic endpoint detection as a \textbf{causal audio-language reasoning task}, where the model determines from partial speech whether the user should retain or yield the conversational floor.

    \item We introduce \textbf{FD-VAD}\footnote{Code and model release:\url{https://fd-vad.github.io/}}, an ASR-free audio-to-LLM architecture that performs streaming semantic endpoint detection for full-duplex speech interaction.
    
    \item We propose \textbf{confidence-gated endpointing} and \textbf{boundary-focused hard-negative sampling} to control the interruption--latency trade-off and improve decisions near true endpoints.

    \item FD-VAD achieves the \textbf{highest EOT recall on TurnBench dev} among compared systems ($0.853$ at $\mathrm{FP}\leq0.10$) in a zero-shot setting, and matches a strong non-streaming turn classifier at the utterance level ($0.965$ vs.\ $0.966$). 
\end{itemize}
\section{Related Work}

\noindent\textbf{Transcription- and segmentation-dependent endpointing.}
Acoustic VAD detects speech boundaries from energy, spectral, or learned cues~\cite{teager,silero}, but provides no measure of semantic completion. Cascaded systems such as TEN~\cite{ten} obtain semantics through an ASR transcript, while Smart Turn~\cite{smartturn} predicts completeness directly from an acoustically segmented utterance. FD-VAD instead operates continuously on causal audio windows without requiring either transcription or a completed speech segment.

\noindent\textbf{Streaming methods with intermediate signals.}
Recent methods reduce this dependence while retaining auxiliary mechanisms. Easy Turn~\cite{easyturn} jointly predicts ASR and turn state, FastTurn~\cite{fastturn} relies on streaming CTC representations, SpeculativeETD~\cite{speculativeetd} triggers semantic inference after detecting silence, and Next-Turn~\cite{nextturn} learns endpoint timing through time-to-next-speech supervision. FD-VAD removes these intermediate pathways and directly maps observed speech to a semantic \textsc{Continue}/\textsc{Stop} state.

\noindent\textbf{Dialogue-coupled and speech-language models.}
Full-duplex models such as Moshi and Freeze-Omni~\cite{moshi,freezeomni} learn turn-taking jointly with response generation, while TurnFSM~\cite{turnfsm} embeds semantic VAD within an LLM-based dialogue-state controller. FD-VAD instead remains independent of the downstream voice agent. Architecturally, it follows the audio encoder--adapter--LLM paradigm used in speech-language models~\cite{llamaomni,slamomni,vocalnet}, specializing it for causal endpoint prediction using bounded audio context and a last-chunk objective.

\begin{figure}[t]
    \centering
    \includegraphics[width=\columnwidth]{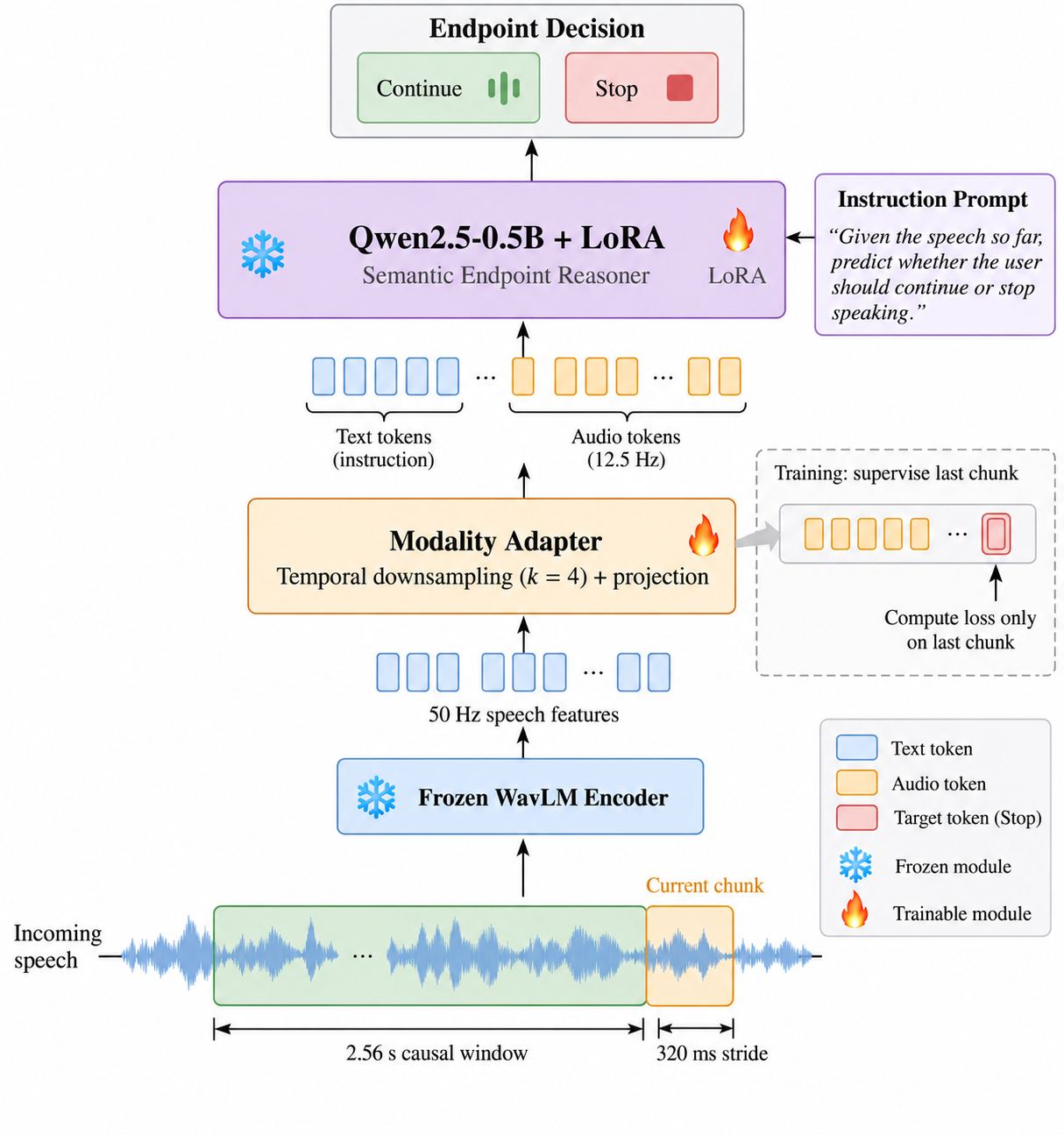}
    \caption{\small{\textbf{FD-VAD architecture.} A causal audio window of incoming speech is encoded by frozen WavLM, projected into the embedding space of a LoRA-adapted Qwen2.5-0.5B language model for \textsc{Continue}/\textsc{Stop} endpoint prediction. Training supervises only the final chunk of each window.}}
    \label{fig:fdvad_architecture}
\end{figure}

\section{Method}
\label{sec:method}


\noindent\textbf{\underline{Architecture}}: Figure \ref{fig:fdvad_architecture} shows FD-VAD, an autoregressive LLM-based architecture for semantic endpoint detection,  containing three components: \textbf{(i) WavLM Audio Encoder.} A frozen WavLM-base-plus encoder~\cite{wavlm} maps raw waveform to frame-level speech features. \textbf{(ii)Audio-Text Modality Adapter.} To bridge the modality gap between audio and text, we utilize a modality adapter composed of two linear layers with a ReLU activation function that temporally downsamples and projects audio features into the embedding space of an LLM. \textbf{(iii) LLM Backbone.} We use Qwen2.5-0.5B-Instruct~\cite{qwen} as the LLM backbone, which takes the adapted audio features concatenated with text prompt embeddings as input to predict the semantic completeness state of the input speech segment. Let $A^I$ denote the WavLM feature sequence. The adapter groups every $k$ consecutive frames and applies $A^{P}=W_2\,\mathrm{ReLU}(W_1A^I+b_1)+b_2$. We use $k{=}4$, so the 50\,Hz speech representation becomes a 12.5\,Hz sequence of audio tokens. The projected audio embeddings are concatenated with a short instruction prompt, and the LLM predicts a single state token, \textsc{Continue} or \textsc{Stop}, followed by \texttt{<eos>}. The speech encoder and base LLM remain frozen. We jointly optimize the modality adapter and low-rank LoRA parameters~\cite{lora}, for 10.3M trainable parameters (1.7\% of the total parameter count). This keeps the endpoint detector modular and parameter-efficient while reusing strong pretrained speech and language representations.


\noindent\textbf{\underline{Causal Sliding-window Inference}}: To enable streaming inference, we employ a sliding window training strategy that makes predictions using only the audio within each window, reducing dependence on the entire input sequence, enabling incremental chunk-wise predictions, and offering latency advantages. FD-VAD operates on a 10\,ms waveform slicing grid. For chunk index $c$, the model advances by a stride of $S_f{=}320$\,ms and retains at most $W_f{=}2.56$\,s of left context. Thus, each decision is made every 320\,ms from a 2.56\,s causal window which is encoded by WavLM into approximately 128 feature frames at 50\,Hz. The adapter downsamples these by $k{=}4$ to approximately 32 audio tokens at 12.5\,Hz before the LLM.


\noindent\textbf{\underline{Last-chunk Training Objective}}: For each window, supervision is applied only to its terminal chunk. Let $y_c\in\{0,1\}$ denote \textsc{Continue} and \textsc{Stop}, respectively, and define the target sequence $Y_c{=}\{y_c,\langle\mathrm{eos}\rangle\}$. We minimize the autoregressive cross-entropy
\begin{equation}
\mathcal{L}(\theta)=-\sum_{t=1}^{|Y_c|}\log p_\theta\!\left(Y_c^{(t)}\mid Y_c^{(<t)},A_c^{P},T^P\right),
\end{equation}
where $A_c^{P}$ denotes the adapted audio embeddings for the current window and $T^P$ is the instruction prompt.


\noindent\textbf{\underline{Confidence Gating Commitment}}: A single erroneous \textsc{Stop} can prematurely interrupt a user. At inference time we therefore optionally commit \textsc{Stop} only when $P(\textsc{Stop})\geq\tau$ for $K$ consecutive decisions. Increasing $\tau$ or $K$ reduces false interruptions but also delays the commit. We treat this confidence gate as an operating-point controller layered on top of the base model.


\noindent\textbf{\underline{Boundary-focused Hard-negative Sampling}}: Training errors concentrate near the true endpoint, where a one-chunk timing shift changes the label. We consider boundary-focused hard-negative sampling that oversamples windows whose terminal chunk lies within $\pm2$ chunks of the endpoint. This leaves inference unchanged but reallocates training mass towards ambiguous boundary cases.

\section{Data and Experimental Setup}
\label{sec:experiment_setting}

\noindent\textbf{\underline{Training Data}:}
We use the English subset of smart-turn-v3.1~\cite{smartturntrain}, a 16\,kHz conversational endpointing corpus with native \emph{complete} and \emph{incomplete} utterance labels. We follow the official train/test partition. The training set contains $63{,}386$ clips ($31{,}338$ complete / $32{,}048$ incomplete; $156$\,h), comprising $42{,}529$ human and $20{,}857$ TTS recordings. For complete utterances, the \textsc{Stop} transition is placed at the acoustic speech offset ($-40$\,dB relative to peak), with subsequent chunks labeled \textsc{Stop}; incomplete utterances remain \textsc{Continue} throughout. Incomplete examples are corpus-native rather than artificially truncated complete utterances.

\noindent\textbf{\underline{Training Hyperparameters}:}
We freeze the WavLM encoder and LLM backbone and train only the modality adapter and LoRA parameters ($r{=}16$) for one epoch using AdamW with learning rate $2\times10^{-4}$, cosine decay, and $3\%$ warmup. Training uses an effective batch size of $64$ in bf16 on two A100 GPUs.

\noindent\textbf{\underline{Evaluation Data and Protocols}:}
The held-out in-domain test set contains $1{,}000$ clips ($500$ complete / $500$ incomplete; $695$ human / $305$ TTS; $2.5$\,h), with median duration $9.1$\,s. For synthetic-to-human transfer, we additionally train on TTS speech only and evaluate on unseen human recordings. We evaluate conversational transfer zero-shot on the TurnBench dev set~\cite{turnbench}, which contains real two-channel human dialogue and uses event-anchored end-of-turn (EOT) evaluation. For this setting, Silero VAD~\cite{silero} provides an acoustic gate and permits an EOT commit after at least $1$\,s of post-speech silence; FD-VAD receives no TurnBench training. We report evaluation results as mean of three random seed runs.

\noindent\textbf{\underline{Evaluation Metrics}:}
At the chunk level, we report precision, recall, $F_1$, and accuracy with \textsc{Stop} as the positive class. We separately report the \emph{chunk false-stop rate}, defined as the fraction of chunks from incomplete utterances incorrectly predicted as \textsc{Stop}. At the utterance level, we report endpoint accuracy separately for complete and incomplete queries, together with $95\%$ bootstrap confidence intervals from $2000$ resamples. We also report strict sentence accuracy (Sent-C/Sent-I), which requires all chunk decisions in an utterance to be correct, and near-boundary $F_1$, computed over predictions within $\pm2$\, chunks of the true endpoint. For confidence-gated endpointing, the \emph{false-interruption rate} is the fraction of incomplete utterances for which the system commits any premature \textsc{Stop}, and endpointing delay is $L=t_{\mathrm{commit}}-t_{\mathrm{EOT}}$, where $L<0$ denotes a commit before the annotated endpoint. We report this delay jointly with false-interruption rate. On TurnBench, we follow its EOT protocol and report recall at $\mathrm{FP}\leq0.10$ together with median commit latency.

\section{Results}
\label{sec:results}

\begin{table}[t]
\centering
\resizebox{0.75\linewidth}{!}{
\setlength{\tabcolsep}{4pt}
\begin{tabular}{lccc}
\toprule
Model & Chunk $F_1$ & Chunk Acc. & False-stop $\downarrow$\\
\midrule
Always-\textsc{Continue} & 0.000 & 0.953 & 0.0\% \\
\midrule
WavLM-base causal~\cite{wavlm}  & 0.183 & 0.721 & 29.1\% \\
WavLM-large causal~\cite{wavlm} & 0.191 & 0.724 & 29.0\% \\
WavLM-large anchor~\cite{wavlm} & 0.218 & 0.787 & 21.6\% \\
Silero VAD~\cite{silero}        & 0.251 & 0.790 & 22.5\% \\
VAP~\cite{vap}                  & 0.279 & 0.803 & 20.1\% \\
\midrule
\rowcolor{bestcyan}
\textbf{FD-VAD} & \textbf{0.832} & \textbf{0.984} & \textbf{0.6}\% \\
\bottomrule
\end{tabular}
}
\caption{\small{Chunk-level semantic endpoint detection on the held-out smart-turn-v3.1~\cite{smartturntest} test set (1,000 utterances; 500 complete and 500 incomplete). \textsc{Stop} is the positive class; false-stop is the fraction of incomplete-speech chunks incorrectly predicted as \textsc{Stop}. FD-VAD outperforms streaming baselines with lowest false-stop rate.}}
\label{tab:chunk}
\end{table}

\begin{table}[t]
\centering
\resizebox{0.85\linewidth}{!}{
\renewcommand{\arraystretch}{1.05}
\setlength{\tabcolsep}{3.2pt}
\begin{tabular}{@{}clcccc@{}}
\toprule
& Model & Str. & Complete & Incomplete & Overall \\
\midrule
\multirow{2}{*}{\rotatebox[origin=c]{90}{\tiny\textsc{\textbf{Acoustic}}}}
& Energy VAD
& \checkmark & \textbf{0.998} & 0.680
& 0.839 {\scriptsize[.82,.86]} \\
& Silero VAD~\cite{silero}
& \checkmark & 0.996 & 0.196
& 0.596 {\scriptsize[.56,.63]} \\

\cmidrule(lr){2-6}

\multirow{6}{*}{\rotatebox[origin=c]{90}{\scriptsize\textsc{\textbf{Streaming}}}}
& VAP~\cite{vap}
& \checkmark & 0.728 & 0.489
& 0.608 {\scriptsize[.58,.64]} \\
& WavLM-base causal~\cite{wavlm}
& \checkmark & 0.824 & 0.344
& 0.584 {\scriptsize[.55,.62]} \\
& WavLM-large causal~\cite{wavlm}
& \checkmark & 0.904 & 0.264
& 0.584 {\scriptsize[.55,.61]} \\
& WavLM-large anchor~\cite{wavlm}
& \checkmark & 0.908 & 0.268
& 0.588 {\scriptsize[.56,.62]} \\
& SoulX-Duplug~\cite{soulx}
& \checkmark & 0.910 & 0.251
& 0.580 {\scriptsize[.55,.61]} \\
& X2-Turn~\cite{x2turn}
& \checkmark & 0.652 & 0.812
& 0.733 {\scriptsize[.71,.76]} \\

\cmidrule(lr){2-6}

\multirow{4}{*}{\rotatebox[origin=c]{90}{\scriptsize\textsc{\textbf{Offline}}}}
& Whisper+LLM
& $\times$ & 0.860 & 0.768
& 0.814 {\scriptsize[.79,.84]} \\
& Whisper+TEN~\cite{ten}
& $\times$ & 0.940 & 0.660
& 0.800 {\scriptsize[.77,.83]} \\
& UltraVAD~\cite{ultravad}
& $\times$ & 0.920 & 0.840
& 0.880 {\scriptsize[.86,.90]} \\
& Smart Turn~\cite{smartturn}
& $\times$ & 0.966 & 0.966
& 0.965 {\scriptsize[.96,.98]} \\

\cmidrule(lr){2-6}

\rowcolor{bestcyan}
& \textbf{FD-VAD}
& \checkmark & 0.972 & \textbf{0.969}
& \textbf{0.976} {\scriptsize[.95,.98]} \\

\bottomrule
\end{tabular}
}
\caption{\small{Utterance-level endpoint classification on the held-out smart-turn-v3.1~\cite{smartturntest} test set with 95\% bootstrap CI over 2K samples. FD-VAD shows the strongest balanced performance across complete and incomplete turns in streaming mode (Str), comparable to the strongest non-streaming semantic classifier.}}
\label{tab:main}
\end{table}

\noindent\textbf{\underline{In-Domain Chunk and Utterance Performance}:}
Tables~\ref{tab:chunk} and~\ref{tab:main} show that FD-VAD reliably detects semantic endpoints while avoiding premature stops. At the chunk level, FD-VAD reaches $F_1{=}0.832$ with only a $0.6\%$ false-stop rate on incomplete utterances. Raw chunk accuracy is less informative because \textsc{Stop} accounts for only $4.8\%$ of the labels; indeed, the always-\textsc{Continue} baseline already achieves $0.953$ accuracy. More importantly, the balanced utterance-level evaluation shows that FD-VAD performs consistently on both complete and incomplete turns ($0.972$/$0.969$), unlike acoustic VAD, whose performance drops sharply on incomplete speech ($0.998\rightarrow0.680$). This gap confirms that acoustic termination alone is insufficient for semantic endpointing. FD-VAD also substantially outperforms the Whisper-based semantic cascades and performs relatively better compared to Smart Turn ($0.965$ vs.\ $0.976$ overall), despite operating streaming rather than on a completed utterances.

\begin{table}[t]
\centering
\resizebox{0.75\linewidth}{!}{
\setlength{\tabcolsep}{4pt}
\begin{tabular}{llccc}
\toprule
Category & System & Recall & FP & Lat.\,(ms) $\downarrow$\\
\midrule
\multirow{65}{*}{Turn-taking}
 & VAP~\cite{vap}                 & 0.841 & 0.045 & \textbf{463} \\
 & ESPnet turn-taking             & 0.836 & 0.074 & 895 \\
 & ESPnet TT (per-ch.)            & 0.640 & 0.100 & 846 \\
 & Smart Turn~\cite{smartturn}    & 0.754 & 0.100 & 1010 \\
 & Moshi~\cite{moshi}             & 0.212 & 0.066 & 771 \\
  & SoulX-Duplug~\cite{soulx} & 0.194 & 0.154$^{\dagger}$ & 89 \\
\midrule
\multirow{4}{*}{Semantic}
 & Kyutai semantic VAD            & 0.803 & 0.100 & 1024 \\
 & Gemini Live                    & 0.665 & 0.047 & 1197 \\
 & OpenAI semantic VAD            & 0.310 & 0.037 & 763 \\
 & OpenAI server VAD  & 0.933 & 0.563$^{\dagger}$  & 281 \\
\midrule
\multirow{5}{*}{Acoustic}
 & WavLM-large anchor             & 0.825 & 0.100 & 1017 \\
 & Mimi endpointer                & 0.759 & 0.047 & 742 \\
 & WavLM-base causal              & 0.485 & 0.100 & 709 \\
 & WavLM-large causal             & 0.472 & 0.100 & 683 \\
 & RMS energy VAD    & 0.595 & 0.547$^{\dagger}$  & $-98$ \\
\midrule
\rowcolor{bestcyan}
\textbf{Ours} & \textbf{FD-VAD}        & \textbf{0.853} & 0.097 & 1019 \\
\midrule
\multicolumn{2}{l}{\emph{Oracle (annotator)}} & 1.000 & 0.000 & 0 \\
\bottomrule
\end{tabular}
}
\caption{\small{TurnBench~\cite{turnbench} dev end-of-turn (EOT) performance under benchmark false-positive constraint ($\mathrm{FP}\leq0.10$). FD-VAD is evaluated zero-shot and achieves the highest recall among qualifying systems. $\dagger$ denotes systems that exceed the false-positive budget.}}
\label{tab:turnbench}
\end{table}

\begin{figure}[t]
\centering
\begin{tikzpicture}

\pgfplotsset{
    basept/.style={
        only marks,
        mark=*,
        mark size=1.35pt,
        color=gray!60
    },
    keypt/.style={
        only marks,
        mark=*,
        mark size=1.8pt,
        color=blue!65!black
    },
    ungatedpt/.style={
        only marks,
        mark=*,
        mark size=2.1pt,
        color=orange!85!red
    },
    ourspt/.style={
        only marks,
        mark=star,
        mark size=3.6pt,
        very thick,
        color=red!80!black
    }
}

\begin{groupplot}[
    group style={
        group size=2 by 1,
        horizontal sep=0.65cm
    },
    width=4.10cm,
    height=4.15cm,
    ymin=0.15,
    ymax=0.95,
    ytick={0.2,0.4,0.6,0.8},
    tick label style={font=\tiny},
    label style={font=\scriptsize},
    title style={font=\scriptsize},
    grid=major,
    grid style={gray!15},
    axis line style={black!75},
    tick style={black!60},
    clip=false
]


\nextgroupplot[
    title={(a) Recall vs.\ false positives},
    xlabel={false-positive rate},
    ylabel={EOT recall},
    xmin=0,
    xmax=0.16,
    xtick={0,0.05,0.10,0.15},
    xticklabels={0,.05,.10,.15}
]

\fill[green!8]
    (axis cs:0,0.15) rectangle (axis cs:0.10,0.95);
\draw[green!45!black,dashed,thin]
    (axis cs:0.10,0.15) -- (axis cs:0.10,0.95);

\addplot[basept] coordinates {
    (0.100,0.640) 
    (0.047,0.665) 
    (0.047,0.759) 
    (0.066,0.212) 
    (0.037,0.310) 
    (0.100,0.754) 
    (0.100,0.485) 
    (0.100,0.472) 
    (0.154,0.194) 
};

\addplot[keypt] coordinates {
    (0.045,0.841) 
    (0.074,0.836) 
    (0.100,0.825) 
    (0.100,0.803) 
};

\addplot[ungatedpt] coordinates {(0.092,0.320)};

\addplot[ourspt] coordinates {(0.097,0.853)};


\draw[black!45,thin]
    (axis cs:0.045,0.841) -- (axis cs:0.025,0.885);
\node[
    font=\tiny,
    fill=white,
    inner sep=0.8pt,
    anchor=south
] at (axis cs:0.025,0.885) {VAP};

\draw[black!45,thin]
    (axis cs:0.074,0.836) -- (axis cs:0.067,0.895);
\node[
    font=\tiny,
    fill=white,
    inner sep=0.8pt,
    anchor=south
] at (axis cs:0.067,0.895) {ESPnet};

\draw[red!65!black,thin]
    (axis cs:0.097,0.853) -- (axis cs:0.118,0.910);
\node[
    font=\tiny,
    text=red!75!black,
    fill=white,
    inner sep=1pt,
    anchor=west
] at (axis cs:0.118,0.910) {\textbf{FD-VAD}};

\draw[black!45,thin]
    (axis cs:0.100,0.825) -- (axis cs:0.126,0.840);
\node[
    font=\tiny,
    fill=white,
    inner sep=0.8pt,
    anchor=west
] at (axis cs:0.126,0.840) {WavLM};

\draw[black!45,thin]
    (axis cs:0.100,0.803) -- (axis cs:0.126,0.790);
\node[
    font=\tiny,
    fill=white,
    inner sep=0.8pt,
    anchor=west
] at (axis cs:0.126,0.790) {Kyutai};

\draw[orange!75!black,thin]
    (axis cs:0.092,0.320) -- (axis cs:0.068,0.275);
\node[
    font=\tiny,
    text=orange!80!black,
    fill=white,
    inner sep=0.8pt,
    anchor=east
] at (axis cs:0.068,0.275) {ungated};

\node[
    font=\tiny,
    text=green!40!black,
    anchor=north east
] at (axis cs:0.098,0.945) {FP$\leq0.10$};


\nextgroupplot[
    title={(b) Recall vs.\ delay},
    xlabel={median commit latency (ms)},
    ylabel={EOT recall},
    xmin=0,
    xmax=1300,
    xtick={0,400,800,1200},
    xticklabels={0,400,800,1200}
]

\addplot[basept] coordinates {
    (846,0.640)
    (1197,0.665)
    (742,0.759)
    (771,0.212)
    (763,0.310)
    (1010,0.754)
    (709,0.485)
    (683,0.472)
    (89,0.194)
};

\addplot[keypt] coordinates {
    (463,0.841)  
    (895,0.836)  
    (1017,0.825) 
    (1024,0.803) 
};

\addplot[ungatedpt] coordinates {(631,0.320)};

\addplot[ourspt] coordinates {(1019,0.853)};

\draw[black!45,thin]
    (axis cs:463,0.841) -- (axis cs:330,0.900);
\node[
    font=\tiny,
    fill=white,
    inner sep=0.8pt,
    anchor=south
] at (axis cs:330,0.900) {VAP};

\draw[red!65!black,thin]
    (axis cs:1019,0.853) -- (axis cs:1130,0.905);
\node[
    font=\tiny,
    text=red!75!black,
    fill=white,
    inner sep=1pt,
    anchor=west
] at (axis cs:1130,0.905) {\textbf{FD-VAD}};

\draw[black!45,thin]
    (axis cs:1017,0.825) -- (axis cs:850,0.775);
\node[
    font=\tiny,
    fill=white,
    inner sep=0.8pt,
    anchor=east
] at (axis cs:850,0.775) {WavLM};

\draw[orange!75!black,thin]
    (axis cs:631,0.320) -- (axis cs:515,0.270);
\node[
    font=\tiny,
    text=orange!80!black,
    fill=white,
    inner sep=0.8pt,
    anchor=east
] at (axis cs:515,0.270) {ungated};

\draw[black!45,thin]
    (axis cs:89,0.194) -- (axis cs:190,0.235);
\node[
    font=\tiny,
    fill=white,
    inner sep=0.8pt,
    anchor=west
] at (axis cs:190,0.235) {SoulX};

\end{groupplot}
\end{tikzpicture}

\caption{\small{TurnBench~\cite{turnbench} dev operating characteristics. Zero-shot gated FD-VAD achieves the highest EOT recall within the benchmark false-positive constraint ($\mathrm{FP}\leq0.10$). The latency view highlights the complementary trade-off between endpoint reliability and commit speed; selected competitors are annotated and remaining baselines are shown in gray.}}
\label{fig:turnbench}
\end{figure}
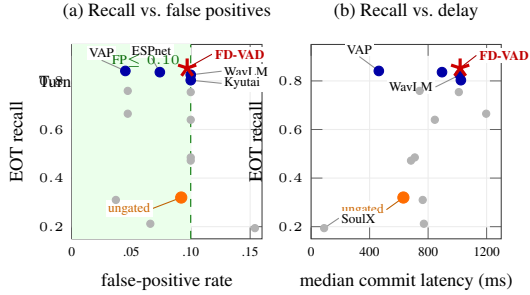

\noindent\textbf{\underline{Zero-Shot Conversational EOT Transfer}:} The in-domain evaluation uses isolated utterances, whereas practical endpointing operates within continuous dialogue. To apply FD-VAD to continuous audio, Silero VAD~\cite{silero} provides an acoustic gate that determines \emph{when} an endpoint is plausible, while FD-VAD determines \emph{whether} the preceding speech is semantically complete. Table~\ref{tab:turnbench} shows that under the TurnBench false-positive constraint ($\mathrm{FP}\leq0.10$), \textbf{FD-VAD achieves the highest EOT recall} among qualifying systems ($0.853$), despite being evaluated zero-shot. However, FD-VAD's $1019$\,ms median commit latency is higher than VAP's $463$\,ms. Fig.~\ref{fig:turnbench} further illustrates this trade-off: FD-VAD lies at the upper edge of the recall--false-positive operating region, but not on the best recall--latency frontier. The large separation between gated and ungated FD-VAD additionally shows that semantic endpoint prediction benefits substantially from acoustic gating in continuous dialogue.

\noindent\textbf{\underline{Streaming Latency Analysis}:} A key advantage of FD-VAD over non-streaming ASR-based endpointing is that semantic inference proceeds concurrently with the user's speech rather than beginning only after the utterance completion. Let $T_{\mathrm{speech}}$ denote the utterance duration, $T_{\mathrm{stride}}$ is FD-VAD's inference stride, and $t_{\mathrm{FD}}$ is it's per-window processing time. An offline ASR-based cascade can produce its semantic endpoint decision only after the utterance has been observed, giving a endpoint decision time of $T_{\mathrm{speech}}+t_{\mathrm{ASR}}+t_{\mathrm{semantic}}$, while FD-VAD continuously updates its endpoint state every $T_{\mathrm{stride}}$($=320$\,ms) while speech is arriving. On a single NVIDIA A100, FD-VAD requires only $45.8$\,ms per $2.56$\,s causal window ($46.8$\,ms p90), and this cost remains constant with utterance duration. At our operating point ($\tau{=}0.9$, $K{=}1$), the resulting median endpoint delay is $190$\,ms with a $2.0\%$ false-interruption rate. Thus, longer utterances increase the latency of non-streaming semantic endpointing, whereas FD-VAD amortizes semantic reasoning through turns.

\begin{table}[t]
\centering
\resizebox{0.95\linewidth}{!}{
\setlength{\tabcolsep}{3.2pt}
\begin{tabular}{lccc|cccc}
\toprule
Variant & Stride & Boundary & Trainable & Chunk & Comp. & Near-b. & Sent-C \\
        & (ms)   & sampling & adapter   & Acc.  & $F_1$ & $F_1$    &   Acc.     \\
\midrule
FD-VAD (base)
& 320 & $\times$ & \checkmark
& 0.981 & 0.827 & 0.838 & 0.406 \\ \midrule

160-ms stride
& 160 & $\times$ & \checkmark
& \cellcolor{green!6}{0.983} & \cellcolor{red!6}{0.798} & \cellcolor{red!6}{0.779} & \cellcolor{red!6}{0.250} \\

Frozen adapter
& 320 & $\times$ & $\times$
& \cellcolor{red!6}{0.978} & \cellcolor{red!6}{0.803} & \cellcolor{red!6}{0.816} & \cellcolor{red!6}{0.378} \\

+ Boundary sampling
& 320 & \checkmark & \checkmark
& \cellcolor{green!6}{\textbf{0.987}} & \cellcolor{green!6}{\textbf{0.838}} & \cellcolor{green!6}{\textbf{0.880}} & \cellcolor{green!6}{\textbf{0.476}} \\
\midrule

Synthetic-only training & 320 & \checkmark & \checkmark & \cellcolor{red!6}{0.972} & \cellcolor{red!20}{0.785} & \cellcolor{red!20}{0.786} & \cellcolor{red!20}{0.341} \\
\bottomrule
\end{tabular}
}
\caption{\small{Ablation results on a matched 8k-clip subset. All variants use the same FD-VAD architecture and differ only in the listed settings. Near-b. is $F_1$ within $\pm2$ chunks of the endpoint; Sent-C is strict all-chunk accuracy on complete utterances. We study robustness to training data by training FD-VAD on synthetic-only data that transfers well per chunk but degrades at utterance-level.}}
\label{tab:abl}
\end{table}

\noindent\textbf{\underline{Ablations and Stability}}: Table~\ref{tab:abl} reveals two useful design trends. First, reducing the stride from $320$\,ms to $160$\,ms slightly raises aggregate chunk accuracy but substantially degrades complete-turn and near-boundary $F_1$. Second, jointly adapting the modality interface and explicitly emphasizing boundary examples are important: freezing the adapter degrades all endpoint-oriented metrics, whereas boundary-focused sampling yields the strongest complete-turn, near-boundary, and strict sentence performance. The three-seed variation is small ($0.9804\pm0.0005$ chunk accuracy), indicating that these trends are not driven by a particular initialization.


\noindent\textbf{\underline{Robustness to Training Data Source}}: FD-VAD model is trained on a mixture of human and synthetic speech. To assess its dependence on human training audio, we additionally train FD-VAD using only synthetic speech and evaluates it on held-out human recordings. The synthetic-only model retains similar per-chunk performance on human speech, with a $1.1$-point reduction relative to mixed-data training, although the stricter all-chunk utterance metrics degrade more substantially, emphasizing the importance of human training data for consistent endpoint decisions across complete utterances.

\section{Conclusion}
We introduced FD-VAD, a streaming semantic endpoint detector that reformulates turn completion as causal audio-language reasoning over partial speech. By mapping bounded audio context directly to \textsc{Continue}/\textsc{Stop} decisions, FD-VAD removes the need for intermediate transcription while remaining modular with respect to the downstream voice agent. Our evaluations show that FD-VAD transfers effectively to continuous real-time dialogue, where semantic reasoning complements conventional acoustic gating. We establish that direct audio-to-language reasoning provides a simple and portable foundation for turn-taking in full-duplex spoken agents.

\clearpage
\newpage
\bibliographystyle{IEEEbib}
\footnotesize{\bibliography{refs}
}

\end{document}